\documentclass[conference]{IEEEtran}
\IEEEoverridecommandlockouts
\renewcommand{\baselinestretch}{0.925}

\usepackage{cite}
\usepackage{amsmath,amssymb,amsfonts}
\usepackage{graphicx}
\usepackage{textcomp}
\usepackage{url}
\usepackage{csquotes}
\usepackage{hyperref}
\usepackage{array}
\usepackage{subcaption}
\usepackage{algorithm}
\usepackage{booktabs}
\usepackage{algpseudocode}
\usepackage{xcolor}
\usepackage{comment}
\usepackage{fancyhdr}
\usepackage[compatibility=false]{caption}

\def\BibTeX{{\rm B\kern-.05em{\sc i\kern-.025em b}\kern-.08em
    T\kern-.1667em\lower.7ex\hbox{E}\kern-.125emX}}

\fancypagestyle{firstpage}{
  \fancyhf{} 

}

\title{Design and Development of a Low-Cost Scalable GSM-IoT Smart Pet Feeder with a Remote Mobile Application}

\author{
\IEEEauthorblockN{
Md. Rakibul Hasan Nishat\textsuperscript{1},
S. M. Khalid Bin Zahid\textsuperscript{2},
Abdul Hasib\textsuperscript{3*},\\
T. M. Mehrab Hasan\textsuperscript{4},
Mohammad Arman\textsuperscript{5},
A. S. M. Ahsanul Sarkar Akib\textsuperscript{6}
}
\IEEEauthorblockA{
\textsuperscript{1,2}Department of Mechatronics Engineering,\\
Rajshahi University of Engineering \& Technology (RUET), Bangladesh\\
\textsuperscript{3*,4,5}Department of Internet of Things and Robotics Engineering,\\
University of Frontier Technology, Bangladesh\\
\textsuperscript{6}Department of Robotics, Robo Tech Valley, Dhaka, Bangladesh\\[4pt]
Emails:
\textsuperscript{1}rakibulnishat1864@gmail.com,
\textsuperscript{2}smkhalidbz@gmail.com,
\textsuperscript{3*}sm.abdulhasib.bd@gmail.com,\\
\textsuperscript{4}mehrabratul210524@gmail.com,
\textsuperscript{5}armaan4421@gmail.com,
\textsuperscript{6}ahsanulakib@gmail.com\\[4pt]
\textsuperscript{*}Corresponding author
}
}

\begin{document}

\maketitle
\thispagestyle{firstpage} 

\begin{abstract}

Pet ownership is increasingly common in modern households, yet maintaining a consistent feeding schedule remains challenging for the owners particularly those who live in cities and have busy lifestyles. This paper presents the design, development, and validation of a low-cost, scalable GSM-IoT smart pet feeder that enables remote monitoring and control through cellular communication. The device combines with an Arduino microcontroller, a SIM800L GSM module for communication, an ultrasonic sensor for real-time food-level assessment, and a servo mechanism for accurate portion dispensing. A dedicated mobile application was developed using MIT App Inventor which allows owners to send feeding commands and receive real-time status updates. Experimental results demonstrate a 98\% SMS command success rate, consistent portion dispensing with $\pm 2.67$\% variance, and reliable autonomous operation. Its modular, energy-efficient design makes it easy to use in a wide range of households, including those with limited resources. This work pushes forward the field of accessible pet care technology by providing a practical, scalable, and completely internet-independent solution for personalized pet feeding. In doing so, it sets a new benchmark for low-cost, GSM-powered automation in smart pet products.

\end{abstract}


\section{Introduction}
Pet care has become an increasingly important concern in modern households, as pets are often regarded as companions and family members. Recent studies shows that more than 60\% US households own a pet. These days, it is usual to see a family owned more than one pet in their house, e.g. cat, dog, or some special pets like a hedgehog, lizard, or even small snake \cite{lee2022petcare,fall}. However, due to busy schedules or unforeseen circumstances, pet owners may have difficulty maintaining a consistent feeding schedule for their furry pet. The concept of automatic pet feeders has gained popularity as a solution to this problem.

In this paper, we proposed a GSM-based advanced pet feeder that ensures automated and reliable feeding for pets. Unlike traditional systems that relies on expensive sensors and complex control mechanisms, our approach uses an Arduino microcontroller as the central controller, a GSM module for remote communication, an ultrasonic sensor to monitor food levels, and a servo motor to control precise portion dispensing. We built a dedicated mobile application, that allows owners to set schedules, send feeding commands and receive real-time notifications via SMS. That ensures consistent care even in the absence of Wi-Fi connectivity. This design provides a cost-effective, user-friendly, and practical solution for modern pet care. Moreover, pets require varying amount of food and minerals based on their size \cite{watson2023drivers}. Recent smart pet feeders have already been advanced from simple timed dispensers to intelligent systems that integrate IoT, AI, and sensor technologies \cite{vijayalakshmi2024design}. Numerous recent designs incorporate cameras, weight sensors, and deep learning algorithms to identify pets and dispense customized servings, while others provide health monitoring via smart collars and real-time behavioral analysis \cite{vijayalakshmi2024innovative}. The primary objective of this project is to bridge the gaps in existing pet feeder technology by designing and fabricating a fully automatic pet food dispenser that combines precise portion control, energy efficiency, and ease of use.

The work has many novelties such as: cost-effective design, scalability, user-friendly interface etc. These features make it suitable for both individual households and large-scale pet care facilities. Furthermore, the integration of IoT allows owners to analyze pet's feeding patterns and adjust schedules accordingly. This also helps promoting pet's healthier lifestyles. The main contributions of this paper are as follows:

\begin{enumerate}
    \item The prototype can do all basic functionalities at an ultra-low total cost of BDT 2500 or USD 35\$. That makes advanced pet care accessible.
    \item The system can operates over GSM networks using SMS. Which completely eliminates the dependence on internet connectivity and makes it functional in remote areas.
    \item It maintains high power efficiency (125 mA idle current) for continuous operation.
\end{enumerate}

\section{Literature Review}
Numerous researchers have developed automated pet feeding systems over the years to ensure that pets are fed on schedule and can be watched over even when their owners are not there. Traditional feeding still depends a lot on human effort, which can be difficult for people with busy or unpredictable schedules. Automated feeders were designed to solve this problem by using programmable timers, IoT connectivity, and remote-control features. But there are still a lot of restrictions, especially when it comes to scalability, cost, adaptability, and dependability in various home settings. These important gaps, which have been found in numerous studies, are directly addressed by GSM Based Pet Feeder.

Babu et al. \cite{babu2019arduino} examined early automated feeding devices that allows owners to feed pets remotely via SMS. While effective in maintaining regular feeding intervals, these systems lacked because it is entirely dependent on strong cellular network connection. Similarly C.W. Huang et al. \cite{10674128} implemented RFID-based feeders that identify pets using electronic tags, allowing controlled meal portions for households with multiple animals. However, these systems were limited by their reliance on tags and a lack of advanced communication features. Furthermore, Amans et al. \cite{amans2025design} designed a precise, scheduled pet feeding system with strong mechanical reliability. But its screw feeder mechanism showed $\pm$10\% error with very small portions and inconsistent food textures.

To overcome accessibility barriers, IoT-enabled feeders were introduced. Zainuddin et al. \cite{zainuddin2024design} designed a Wi-Fi-based pet feeder allowing users to monitor feeding schedules remotely via mobile app. Similarly, Al-Hafiz et al. \cite{al2024design} proposed an IoT-based automated feeder for cats through a smartphone application. Despite the fact that both studies showed how well the system improved animal welfare, its reliance on constant Internet connectivity limited its dependability in areas with weak networks. Sujatha et al. \cite{10841084} explored both Bluetooth-enabled and Wi-Fi-based feeders, offering short-range control and energy efficiency.

Tangka et al. \cite{tangka2021development} developed a poultry farming IoT-based system that supports sustainability and small-to medium-sized farm's financial viability by automating feeding, cutting waste, and guaranteeing a steady supply of food. Its scalability in resource-constrained environments is, however, limited by its high initial costs, technical complexity, and dependence on reliable power and connectivity. Razali et al. \cite{razali2021smart} presented an automated pet feeding system with remote monitoring and predictive analytics to anticipate food shortages and optimize budgeting. The main drawback of this prediction model is that, the system relies heavily on historical data. In addition, the system relies on stable internet connectivity for notifications and data processing.

Airikala et al. \cite{airikala2021automatic} integrated solar energy with IoT-based pet feeding systems that can provide sustainable autonomous feeding even in remote or power-limited areas. But, the limitations include dependence on solar availability and battery capacity, which restrict uninterrupted operation during prolonged cloudy periods.

Additionally, Vijayalakshmi et al. \cite{10690348} studied AI-powered pet care systems that improve home security and pet well-being by combining automated feeding, behavioral analysis, and health monitoring.  Despite these benefits, their practical applications are limited by their high cost, dependence on multiple technologies, and privacy issues. Similarly, Kotwal et al. \cite{kotwaldeep} proposed an AI-powered system that enables a precise, behavior-based automated feeding system. Here, the system's complexity, reliance on multiple sensors and stable Wi-Fi, and high computational demands made it expensive and less accessible for the average user. Saadoon et al. \cite{saadoon2023intelligence} introduced an AI-driven system for feeding stray cats using real-time detection. However, its dependence on specialized hardware\cite{11077519} (Jetson Nano, GSM/GPS modules) and sensitivity to environmental and network conditions increased cost and complexity which potentially limited consistency in practical deployments.

\section{System Design \& Architecture}
\begin{figure}[h]
    \centering
    \includegraphics[width=0.95\linewidth]{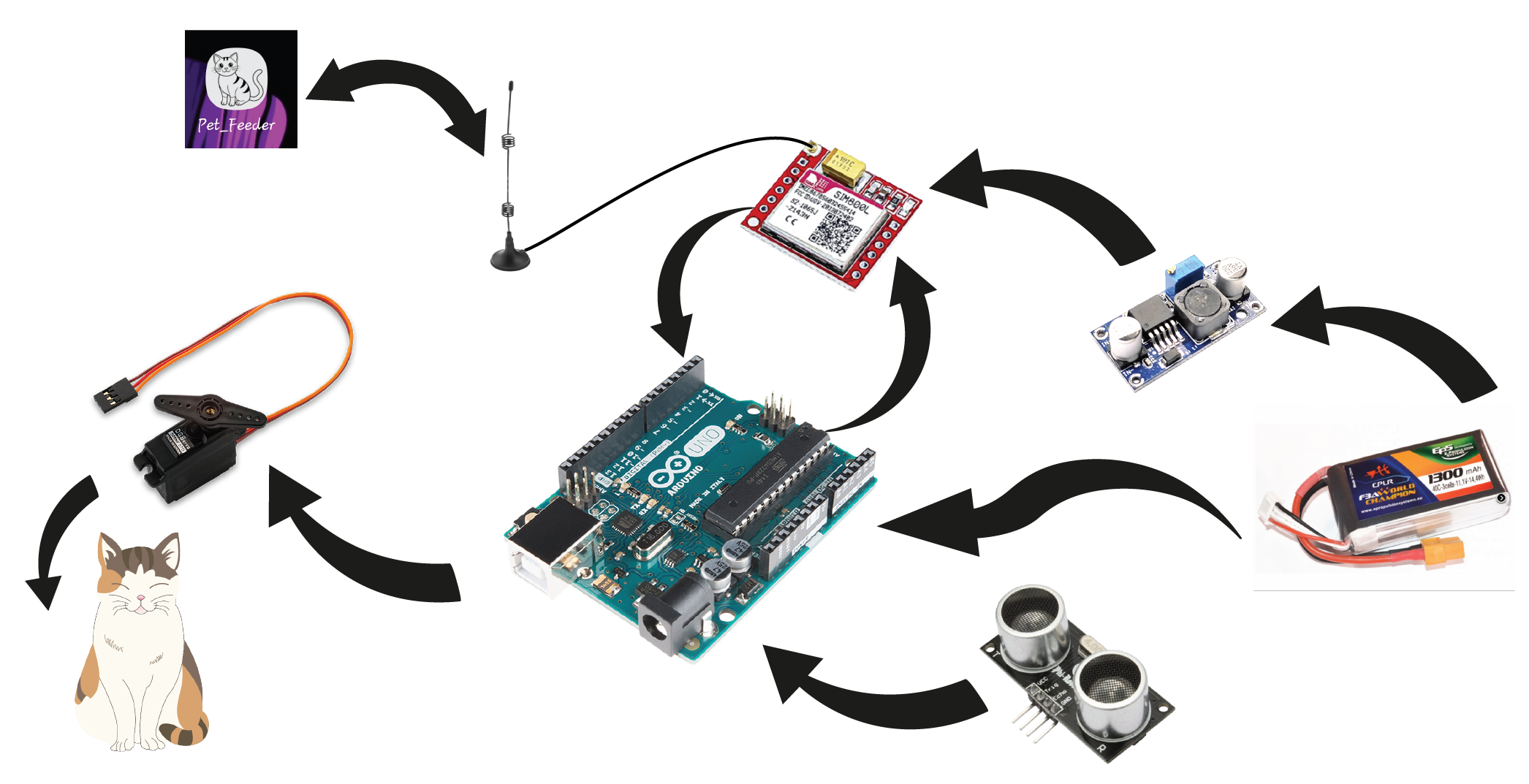}
    \caption{Proposed System Diagram for Controlling each Components}
    \label{fig:placeholderr}
\end{figure}

The proposed system is designed to make a robust, cost-effective, and user-friendly solution for remote pet feeding. The design consists of two main subsystems: software, which controls communication, user commands, and control logic, and hardware, which performs physical actuation and sensing. Fig. \ref{fig:placeholderr} illustrates the system design structure.

\subsection{Hardware Architecture}

\begin{table}[h]
\scriptsize
\centering
\caption{Hardware Components and Specifications}
\label{tab:hardware}
\begin{tabular}{|p{1.7cm}|p{2.3cm}|p{3.5cm}|}
\hline
\textbf{Component} & \textbf{Functionality} & \textbf{Specifications} \\ \hline
Arduino Uno\cite{11160079,edubot} & Central control unit & ATmega328P, 16 MHz, 32 KB Flash, 2 KB SRAM \\ \hline
GSM Module (SIM800L)\cite{pico2025internet} & Communication  & GSM/GPRS (850/900/1800/1900 MHz), 3.4V - 4.4V \\ \hline
Servo Motors\cite{11141132,bionic} & Gesture articulation & 7 units, 4.8-6V, 180° rotation, 2.5 kg-cm torque \\ \hline
Ultrasonic Sensor\cite{Mussa2025} & Measurement & 2 - 400 cm range, 30 degree, +5V DC, 15mA   \\ \hline
Buck Converter & Voltage Regulator & 4.5V-8V DC Input, 0-1.6A, 1.25MHz \\ \hline
Lipo Battery & Power Source & 12V and 1600mAh \\ \hline
\end{tabular}
\end{table}

\begin{figure}[h]
    \centering
    \includegraphics[width=0.95\linewidth]{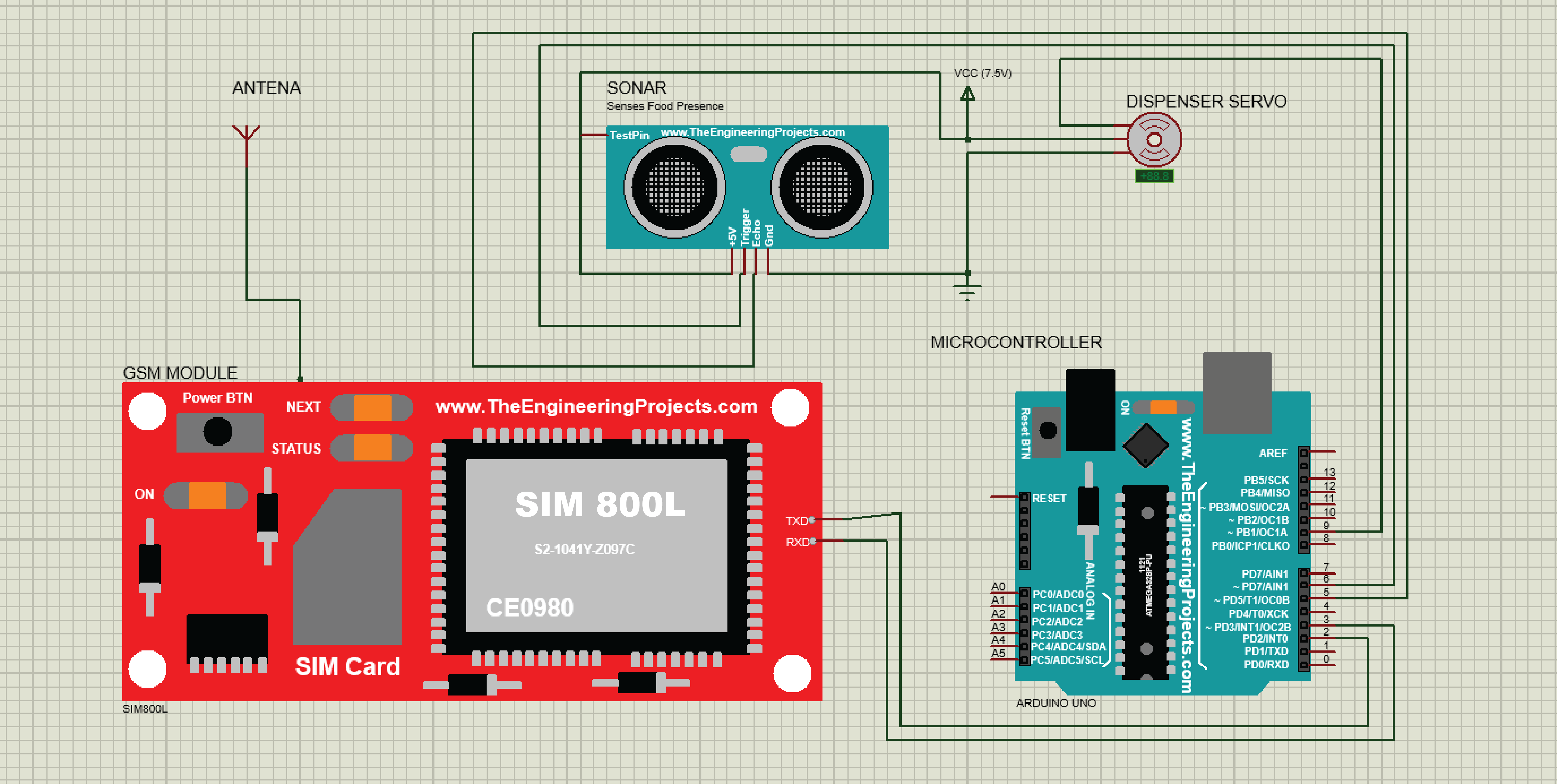}
    \caption{Schametic Diagram of the Project}
    \label{fig:placeholderrr}
\end{figure}

The key hardware components are listed in Table \ref{tab:hardware}. The system architecture schematic is presented in Fig. \ref{fig:placeholderrr}. The Arduino Uno was selected as the central microcontroller for its simplicity, low idle power consumption, and cost-effectiveness. The SIM800L GSM module was chosen to provide internet-independent connectivity via ubiquitous cellular networks, a core design requirement. An HC-SR04 ultrasonic sensor offered a low-cost, non-contact method for food-level monitoring, sufficient for this application. Finally, the SG90 servo motor was selected for its precise angular control and the ability to actuate the dispensing gate reliably with minimal power draw.

\subsection{Mobile Application Development} 

\begin{figure}[h]
    \centering
    \begin{subfigure}{0.165\textwidth}
        \centering
        \includegraphics[width=\linewidth]{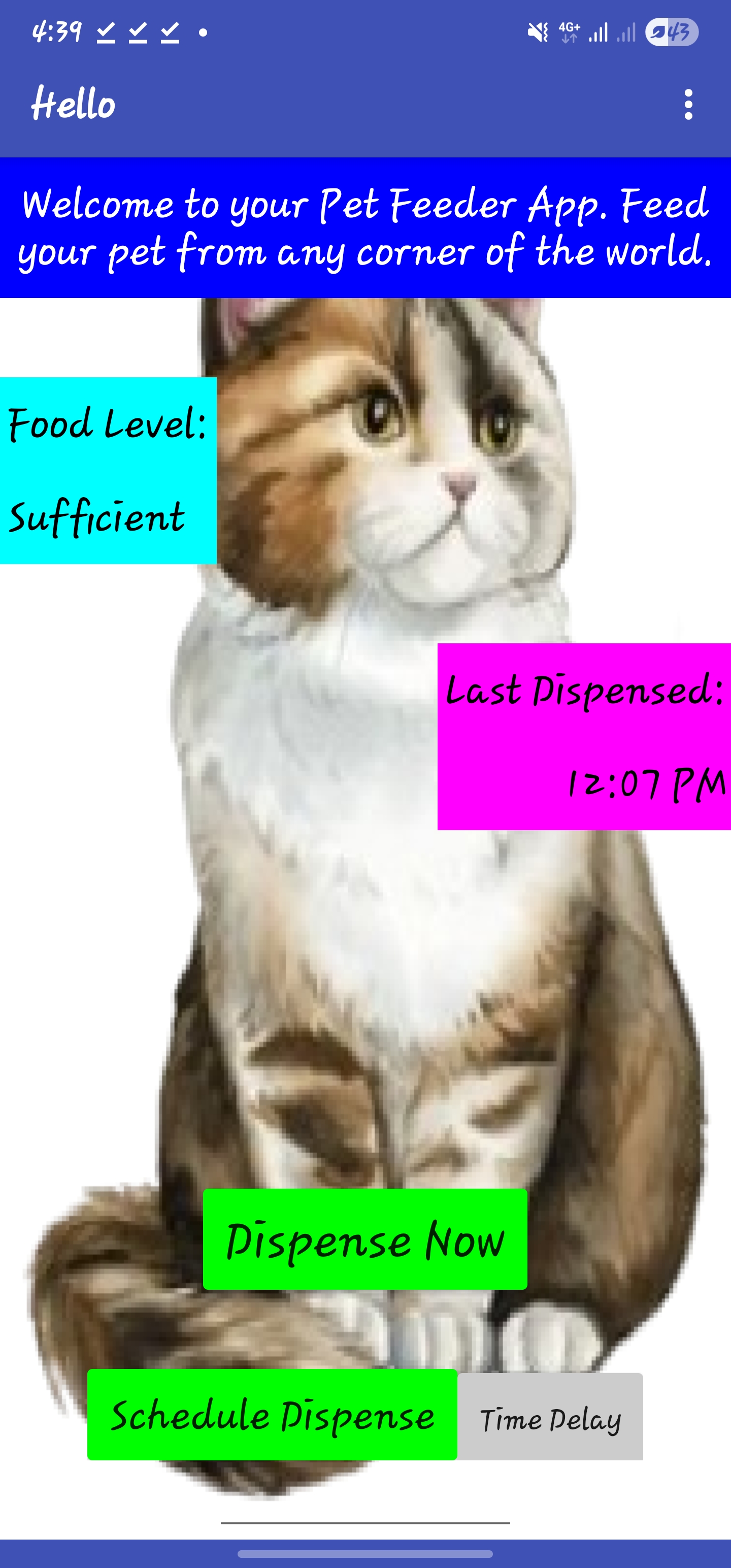}
        \caption{ Mobile Application Interface}
        \label{fig:app_interface}
    \end{subfigure}
    \hfill
    \begin{subfigure}{0.28\textwidth}
        \centering
        \includegraphics[width=\linewidth]{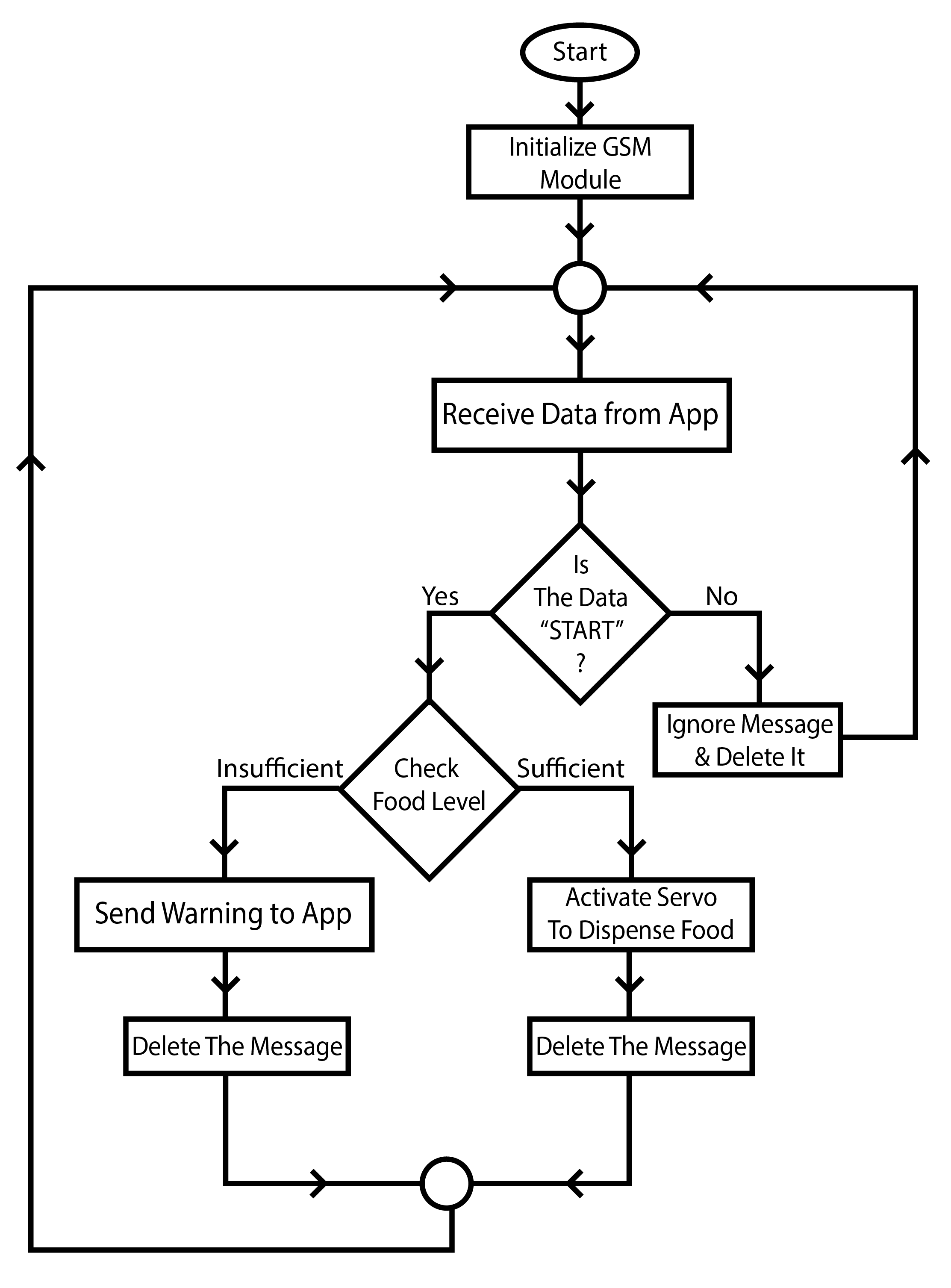}
    \caption{Workflow Diagram of Proposed System}
    \label{fig:ppplaceholder}
    \end{subfigure}
    
    \caption{(a) Mobile Application Interface, (b) Workflow Diagram}
    \label{fig:app_collage}
\end{figure}

The mobile app is built using the MIT App Inventor. MIT App Inventor offers a visual, block-based programming interface, that allows to develop an app without having prior knowledge in Programming Languages. It is open sourced, well documented and supports cloud based projects. And it's free to use. The developed App interface is shown in Fig. \ref{fig:app_interface}.

\subsection{System Workflow}

The system operates through a structured sequence of events, to ensure reliable and autonomous pet feeding as shows in Fig. \ref{fig:ppplaceholder}

\subsubsection{Control Equations}

\begin{equation}
\theta = \frac{\text{PWM}_{\text{signal}}}{255} \times 180^{\circ}
\end{equation}

\begin{equation}
d = \frac{t \times v_{\text{sound}}}{2}
\end{equation}

$\theta$ = The resulting angular position of the servo horn in degrees.

$d$ = The calculated distance to the target.

$t$ = The time elapsed between the emission of the ultrasonic pulse and the reception of its echo

The smart pet feeding system follows the Algorithm\ref{alg:petfeedeer}:

\begin{algorithm}
\scriptsize
\caption{GSM-Based Smart Pet Feeder Protocol}
\label{alg:petfeedeer}
\begin{algorithmic}[1]
\State \textbf{Initialize:} $\{M$ (Arduino), $G$ (SIM800L), $U$ (HC-SR04), 
\Statex \hskip2em $S$ (Servo), $N$ (Authorized Numbers), $T$ (Scheduler)\}
\State Configure serial for GSM; attach servo in CLOSED position
\State Load authorized phone numbers from EEPROM
\State Setup timers for scheduled feeding and periodic food checks
\While{System Power $P = 1$}
    \Comment{Main Loop}
    \Statex
    \State \textbf{Mode 1: Response to Incoming SMS}
    \If{$G.\texttt{available}()$}
        \State $rawSMS \gets G.\texttt{readMessage}()$
        \State $sender \gets \texttt{parseNumber}(rawSMS)$
        \State $cmd \gets \texttt{parseCommand}(rawSMS)$
        \If{$\texttt{isAuthorized}(sender)$}
            \If{$\texttt{validatePIN}(cmd)$}
                \State $\texttt{executeCommand}(cmd)$ \Comment{FEED, STATUS, RESET}
            \Else
                \State $G.\texttt{sendSMS}(sender, "ERROR: Invalid PIN")$
            \EndIf
        \Else
            \State $G.\texttt{sendSMS}(sender, "ERROR: Unauthorized Number")$
        \EndIf
    \EndIf
    \Statex
    \State \textbf{Mode 2: Scheduled Feeding}
    \If{$\texttt{currentTime} = T.\texttt{nextSchedule}$}
        \State $\theta \gets \frac{PWM_{signal}}{255} \times 180^\circ$ \Comment{Servo angle, Eq.~(1)}
        \State $S \gets \theta$ 
        \State $\texttt{logEvent}(\text{"Scheduled Feed"})$
    \EndIf
    \Statex
    \State \textbf{Mode 3: Periodic Food Level Check}
    \If{$\texttt{foodCheckTimer.elapsed}()$}
        \State $t \gets U.\texttt{measureEchoTime}()$
        \State $d \gets \frac{t \times v_{sound}}{2}$
        \If{$d > d_{threshold}$}  \Comment{Distance, Eq.~(2)}
            \State $G.\texttt{broadcastSMS}("ALERT: Low Food Level")$
        \EndIf
    \EndIf
\EndWhile
\end{algorithmic}
\end{algorithm}

\subsection{Cost Analysis}

Table \ref{tab:hardwaree} shows the detailed Bill of Materials (BOM), demonstrating that the entire system can be fabricated for approximately 2500 BDT or \$35. This cost is a fraction of that of commercial smart feeders, which often retail for well over \$100, and is significantly lower than more complex AI- or Wi-Fi-dependent systems reported in the literature \cite{10690348}, \cite{kotwaldeep}.

\begin{table}[h]
\scriptsize
\centering
\caption{Detailed Bill of Materials}
\label{tab:hardwaree}
\begin{tabular}{|p{2cm}|p{1cm}|p{1cm}|p{1cm}|}
\hline
\textbf{Component} & \textbf{Quantity} & \textbf{Cost (BDT)} & \textbf{Cost (USD)} \\ \hline

Arduino Uno & 1 & 380 & 4 \\ \hline

GSM Module (SIM800L) & 1  & 350 & 8 \\ \hline
Servo Motors & 1 & 150  & 4 \\ \hline
Ultrasonic Sensor & 1 & 100 &  2 \\ \hline

Buck Converter & 1 & 100 & 2 \\ \hline

Lipo Charger & 1 & 320 & 4 \\ \hline

PVC Board & 1 & 100 & 2 \\ \hline 

Lipo Battery & 1 & 1000 & 9 \\ \hline 

Total Cost & - & 2500 & 35 \\ \hline

\end{tabular}
\end{table}

\subsection{Security Design and Threat Mitigation}
The proposed system is vulnerable to a number of possible security risks, such as unauthorized access, and SMS spoofing, Erroneous Commands from unknown numbers because it depends on GSM-based communication. To prevent this, authorized phone numbers are stored in EEPROM, and every command must include a valid PIN (e.g., FEED) before execution. It provides an essential first layer of defense against casual misuse and erroneous commands. Input parsing ensures that only properly formatted commands are accepted, reducing the risk of spoofed SMS.

\section{Implementation \& Prototype}

\subsection{Hardware Integration}
The hardware assembly process involved systematic integration of the core components as shown in Fig. \ref{fig:placeholderr}. Each component was first individually checked by their functionally and then validated to ensure optimal performance before integration.

\subsection{Mobile Application Integration}

The mobile application was built using \enquote{MIT App Inventor} website, a visual programming tool that makes rapid development easier. During the integration process, the main software components (listed in Fig. \ref{fig:app_collage}) were added step by step. Each part was tested separately to make sure that it worked correctly and communicated smoothly with the hardware controller before integrating everything into the full system.

\subsection{Physical Assembly} 
\begin{figure}[h]
    \centering
    \includegraphics[width=0.95\linewidth]{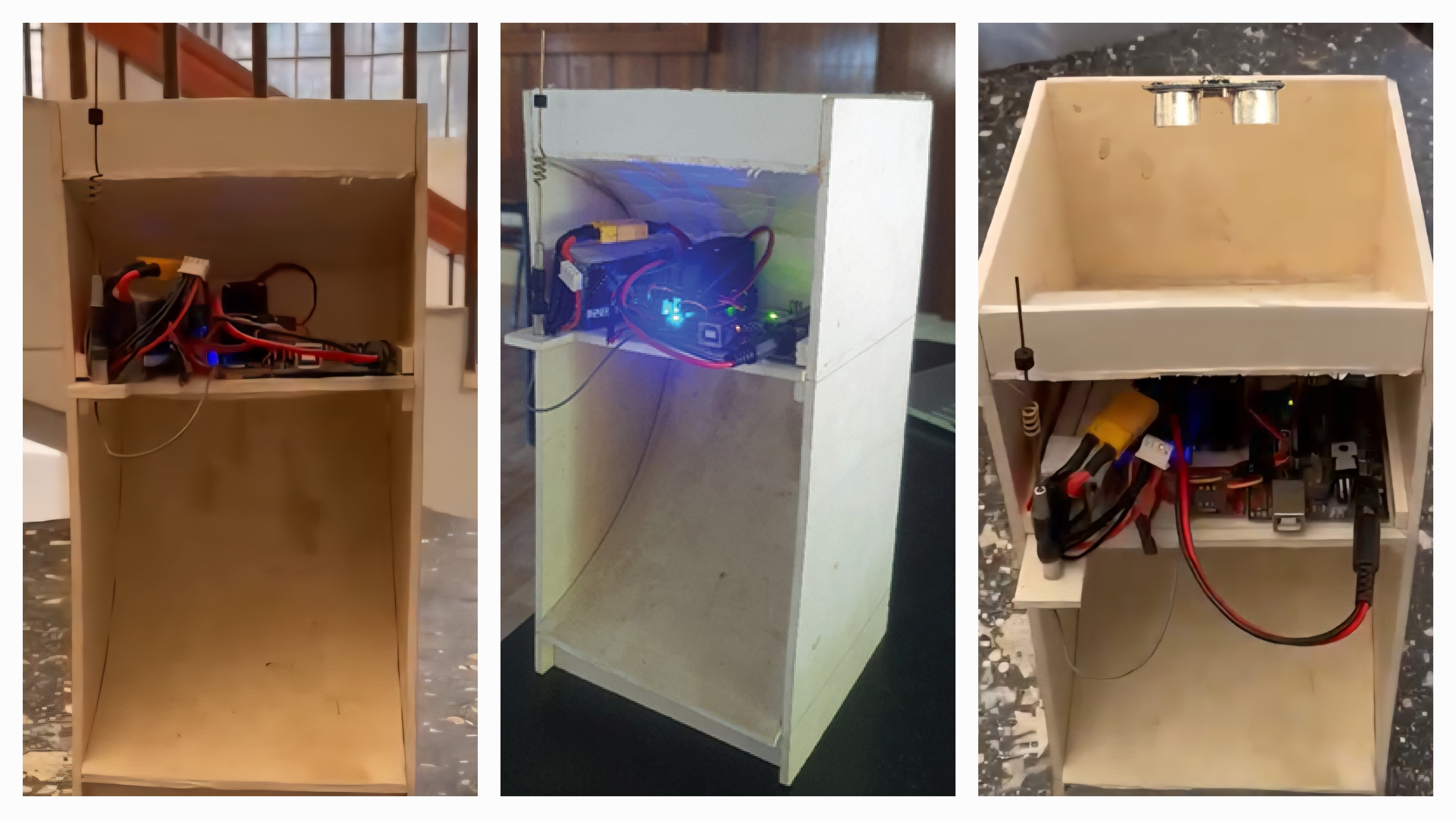}
    \caption{Smart Pet Feeder Prototype}
    \label{fig:placeholderrrr}
\end{figure}

The physical assembly used a modular, PVC board structure to systematically integrate electronic and mechanical components. Each component was secured and wired before final integration and testing Fig. \ref{fig:placeholderrrr}

\section{Experimental Results \& Performance Analysis}

\subsection{Ultrasonic Sensor Reliability}

\begin{figure}
    \centering
    \includegraphics[width=1\linewidth]{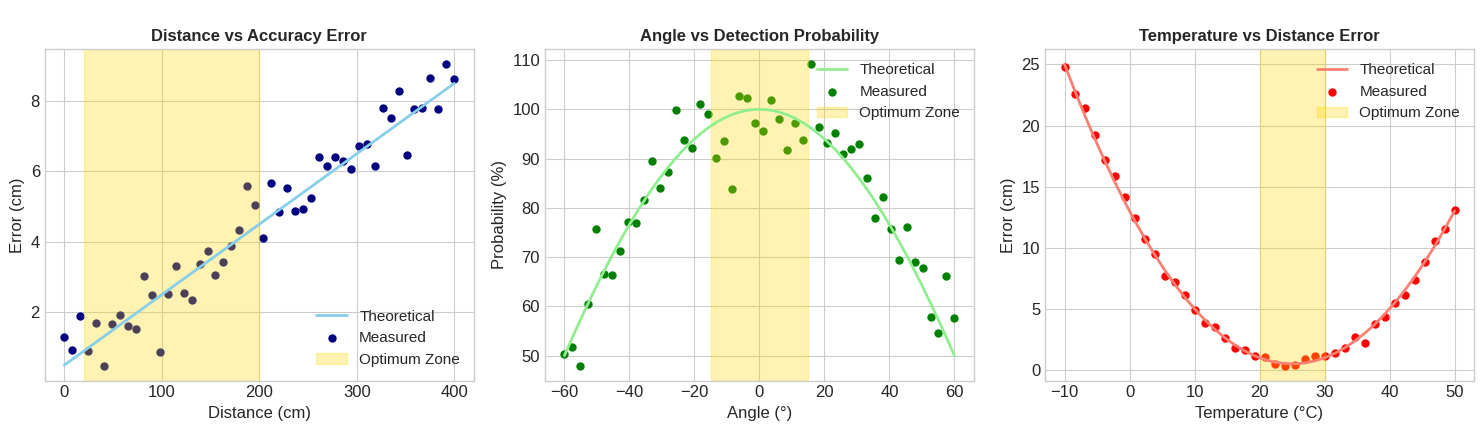}
    \caption{Ultrasonic Sensor Performance Analysis (Theoretical vs Actual)}
    \label{fig:pllaceholder}
\end{figure}

The performance analysis of the ultrasonic sensor confirms that, it is highly reliable within the optimal operating range of 10 to 200 cm. The accuracy decreases for longer ranges. The detection probability is highest near $0^\circ$ and falls significantly beyond $\pm 20^\circ$. This highlights the importance of proper sensor allignment. Temperature test shows that the minimum error is around 20 to $30^\circ$C. Fig. \ref{fig:pllaceholder} demonstrated the ultrasonic sensor's performance analysis.

\subsection{GSM Command Reception Accuracy}

\begin{figure}[h]
    \centering
    \begin{subfigure}{0.24\textwidth}
        \centering
        \includegraphics[width=\linewidth]{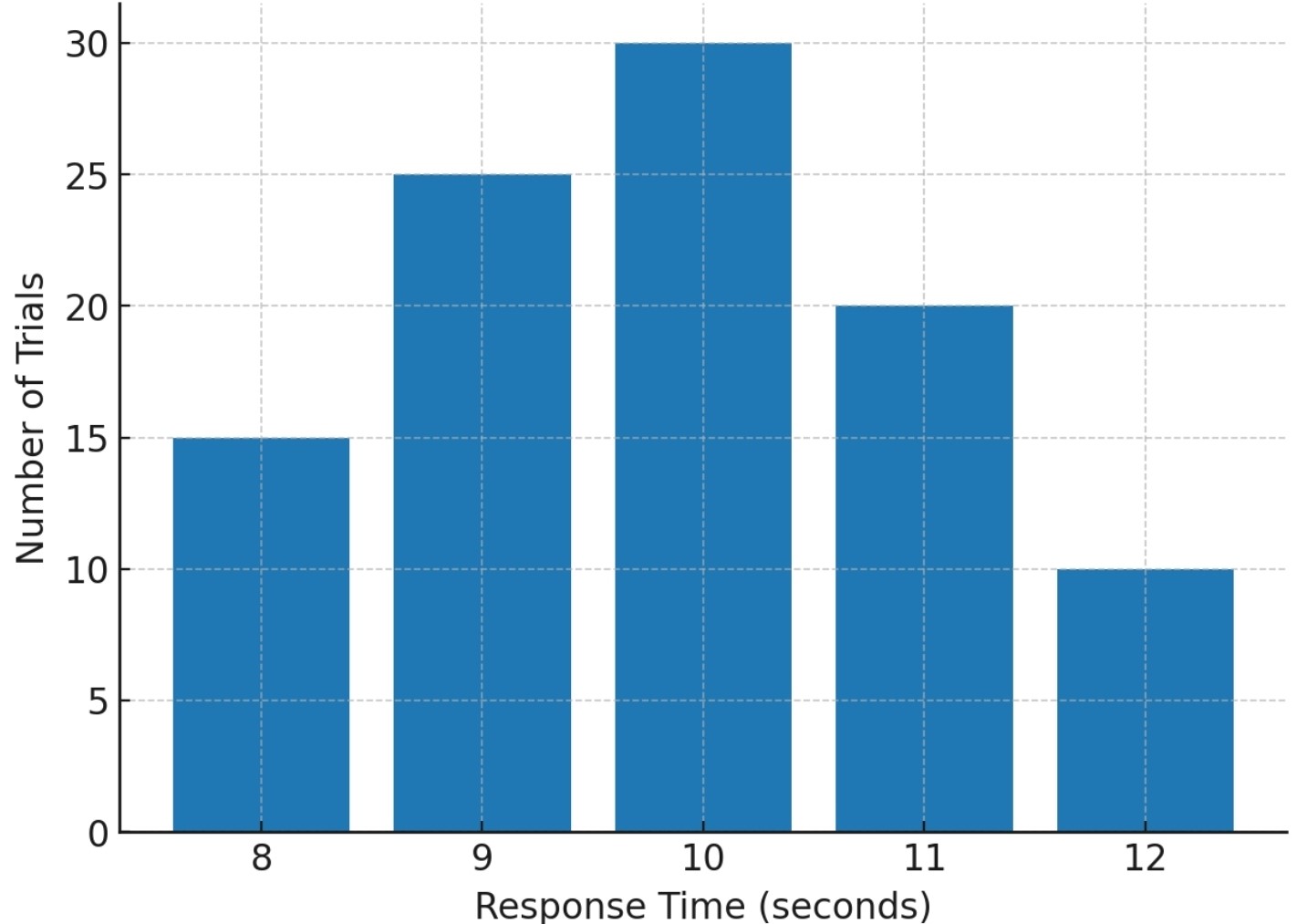}
        \caption{System Response Time Distribution}
        \label{fig:response_time}
    \end{subfigure}
    \hfill
    \begin{subfigure}{0.24\textwidth}
        \centering
        \includegraphics[width=\linewidth]{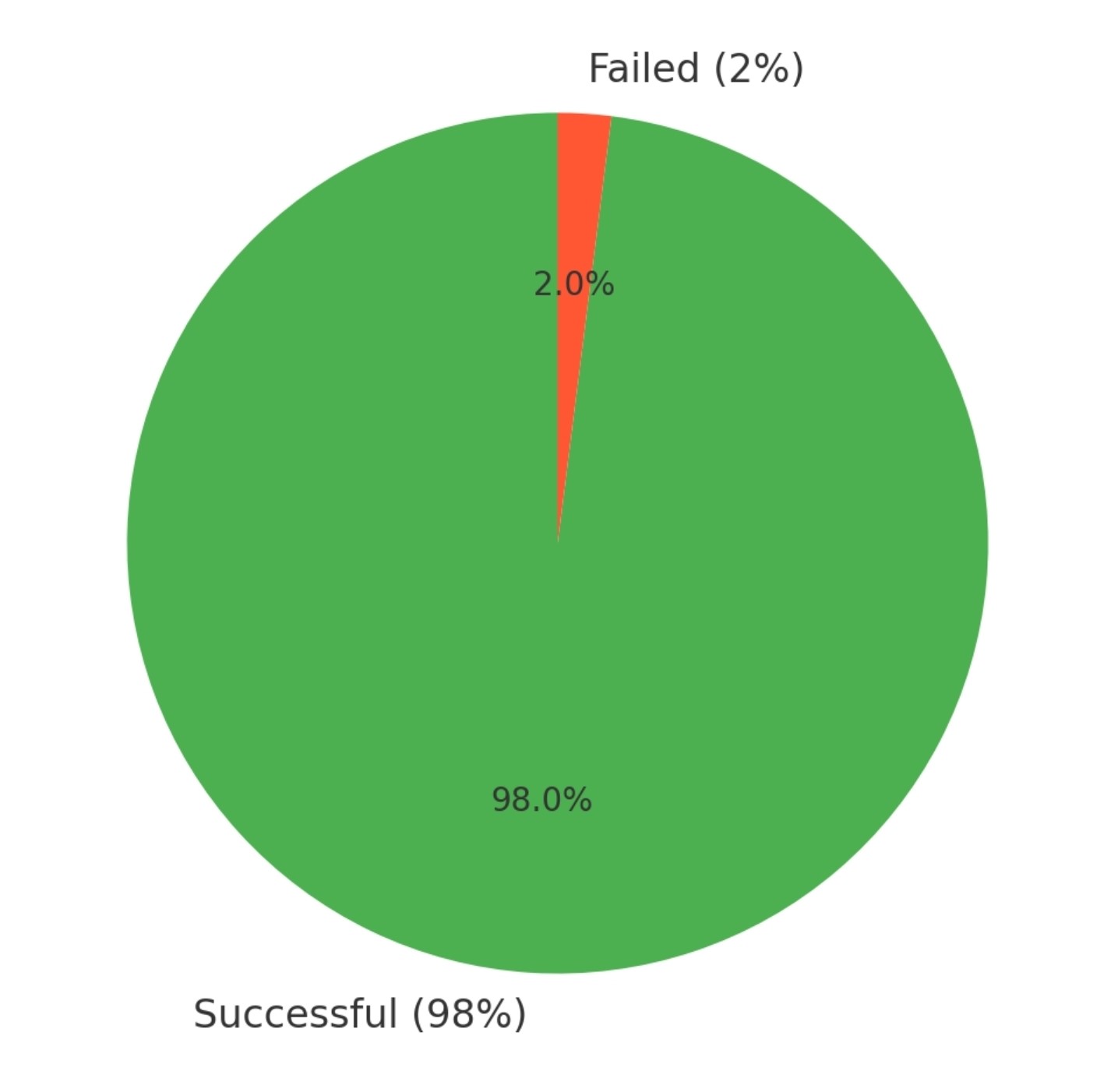}
        \caption{Successful Execution Rate}
        \label{fig:success_rate}
    \end{subfigure}
    \caption{Performance evaluation: (a) System response time distribution, (b) SMS command success rate}
    \label{fig:performance}
\end{figure}

Our system performed reliably when tested using remote commands. Over 100 SMS commands were sent from various locations and network providers during the trials, and approximately 98\% of them were successfully received and executed by the system. On average, it took between 8-12 seconds from sending the ``FEED'' command to execute the feeding sequence, as shown in Fig. \ref{fig:performance}. The delay was mainly for cellular network latency.

\subsection{Performance Analysis of Servo Motor}

\begin{figure}[h]
    \centering
    \includegraphics[width=.9\linewidth]{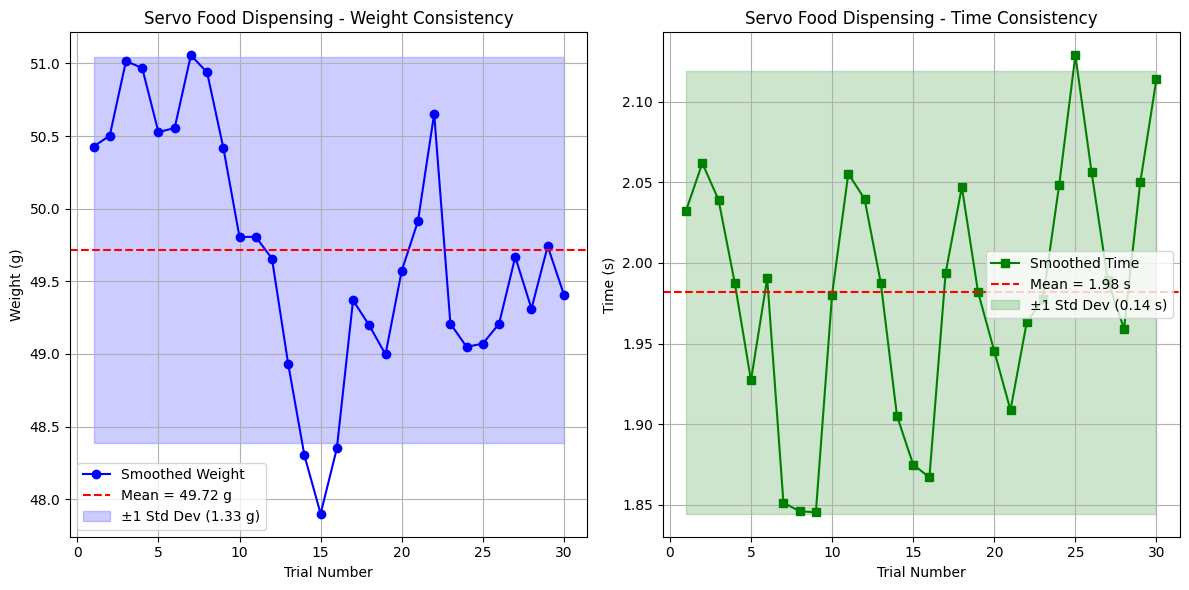}
    \caption{Performance Evaluation of Servo Motor}
    \label{fig:placehoolder}
\end{figure}

The servo dispensing performance graphs shows in Fig. \ref{fig:placehoolder} that the system's consistency in both the volume of food distributed and the time required per operation over thirty trials. The natural fluctuations observed around the mean are expected and reflect normal mechanical and material variations during real-world use. The weight distribution graph shows that the servo consistently dispensed close to 50 grams per cycle, with the shaded region representing the standard deviation, confirming that most trials fell within a narrow and acceptable range. The time graph also shows that the dispensing time averaged about 2 seconds, with minor fluctuations from trial to trial because of variations in load and servo reaction time. The system also demonstrated a $\pm 2.67$\%\% consistency in portion size.

\subsection{Power Consumption Profile}

\begin{figure}[h]
    \centering
    \includegraphics[width=.9\linewidth]{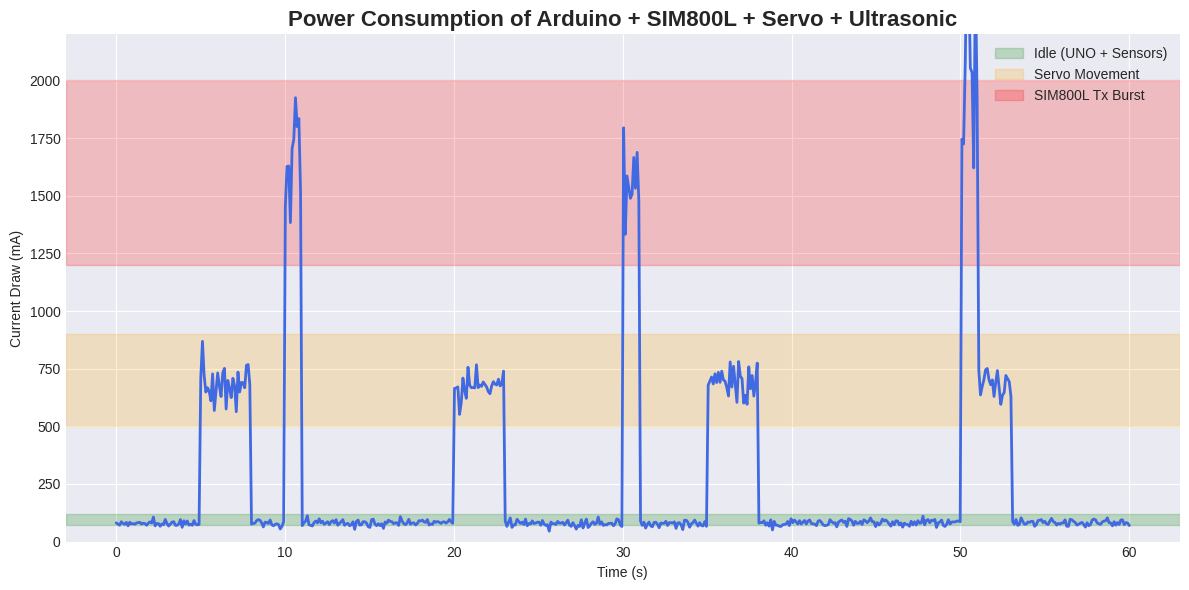}
    \caption{Power Consumption Profile}
    \label{fig:plllaceholder}
\end{figure}

Power consumption of our prototype varied for time to time due to some factors. The factors and their effect on power consumption has been given in Fig. \ref{fig:plllaceholder}.
The power usage graph gives a realistic picture of how the Arduino UNO, SIM800L, ultrasonic sensor, and servo motor work together when the system is powered by a 12V LiPo battery.  Because it operates continually to control the system, the Arduino UNO sets the baseline power draw and keeps it constant.  As is common with GSM modules, the SIM800L causes abrupt increases in consumption during communication bursts, especially when send or receive SMS, make calls, or send data over GPRS. The servo motor's power consumption varies with movement and applied load; it consumes very little in idle conditions but produces considerable peaks with each dispensing or turning operation.  With only a slight, steady load added during distance measurement cycles, the ultrasonic sensor makes a negligible contribution in comparison to the other parts.  These patterns indicate the dynamic nature of system's power usage. Here the baseline consumption is constant and the abrupt brief spikes are introduced during servo operations and SIM800L transmission.  This highlights how crucial it is to have a steady power supply and appropriate control to keep the system dependable even during high current demand.

\subsection{System Testing \& Reliability}

\begin{figure}[h]
    \centering
    \includegraphics[width=.9\linewidth]{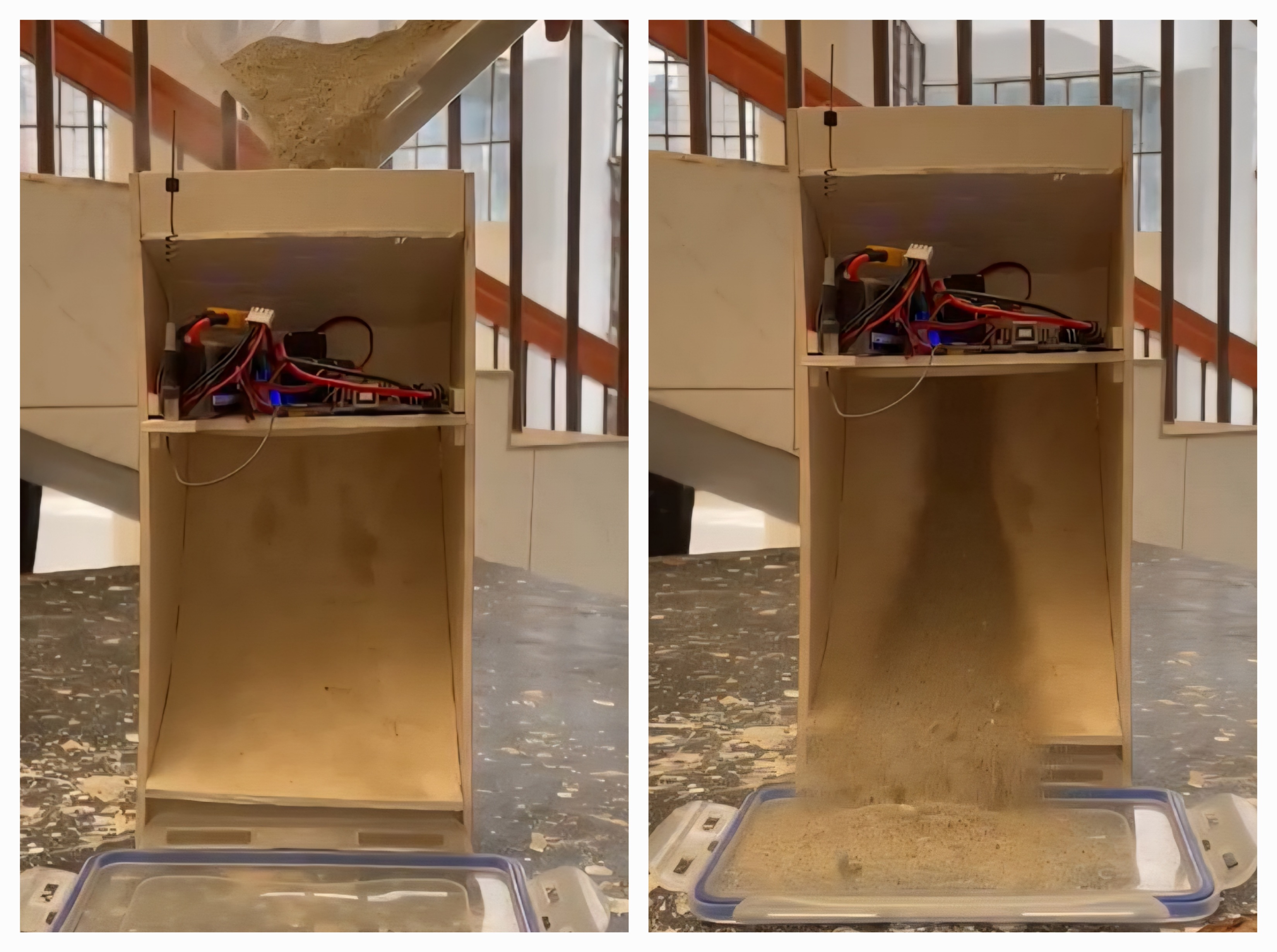}
    \caption{Loading and Dispensing of Food}
    \label{fig:pllllaceholder}
\end{figure}
The prototype was tested for evaluation and it performed very well and met all our requirements. In Fig. \ref{fig:pllllaceholder} the first image shows us the process of loading food in the prototype's hover and the second one shows how the food is dispensed by the prototype. The complete prototype was operated continuously for a month, with automatic schedules activated that is three times daily. During this automatic testing time, no communication failures or missed feeding event was documented. The system also effectively produced low-food notifications when the container threshold level was reached which shows its reliable end-to-end functionality.

\subsection{Comparative Analysis of Existing Systems}

As evidenced in Table \ref{tab:comparative_analysis}, a clear gap exists in the landscape of automated feeders. While solutions offering advanced features like AI-powered pet identification \cite{kotwaldeep} exist, they are cost-prohibitive and reliant on stable Wi-Fi. Conversely, cost-effective solutions either lack remote connectivity and monitoring entirely \cite{amans2025design} or, like \cite{babu2019arduino}, offer only basic SMS commands without automated scheduling or status updates. functionality.

\begin{table}[h]
\centering
\caption{Comparison of Existing Automated Pet Feeder Systems}
\label{tab:comparative_analysis}
\begin{tabular}{p{1cm} *{6}{>{\centering\arraybackslash}p{0.85cm}}}
\toprule
\textbf{Study} & \textbf{\cite{babu2019arduino}} & \textbf{\cite{amans2025design}} & \textbf{\cite{zainuddin2024design}} & \textbf{\cite{al2024design}} & \textbf{\cite{kotwaldeep}} & \textbf{Our System} \\
\midrule

Connectivity & GSM & None & Wi-Fi & Wi-Fi & Wi-Fi & \textbf{GSM} \\
\midrule
Internet-Dependent & \checkmark & \checkmark & \checkmark & \checkmark & \checkmark & \textbf{\texttimes} \\
\midrule
Scheduled Feeding & \texttimes & \checkmark & \checkmark & \checkmark & \checkmark & \checkmark \\
\midrule
Real-time Alerts & \texttimes & \texttimes & \checkmark & \checkmark & \checkmark & \checkmark \\
\midrule
Food-Level Monitoring & \texttimes & \texttimes & \texttimes & \texttimes & \checkmark & \checkmark \\
\midrule
Estimated Cost & Low & Medium & High & High & Very High & \textbf{Very Low} \\
\midrule
Power Source & Mains & Mains & Mains & Mains & Mains & Mains \\
\midrule

\end{tabular}
\end{table}

The GSM-based pet feeder effectively mitigates these collaborative limitations of existing feeding systems that rely heavily on Wi-Fi or advanced AI technologies as they may not be reliable or accessible in all environments. Compared to expensive AI- or IoT-based systems, GSM provides a more affordable, energy-efficient, and user-friendly solution. The unique aspect of this work is the combination of its features: it bridges the identified gaps in affordability, connectivity, and functionality by combining the automated scheduling and status monitoring of high-end IoT systems with the internet-independent reliability of GSM at a very low cost.

\section{Discussion, Limitations \& Future Work}
According to the experimental findings, the suggested GSM-IoT-based pet feeder effectively satisfies its design goals of dependability, affordability, and ease of use. With a 98\% command success rate, the system proves the reliability on SMS-based communication systems which makes it particularly valuable in regions where the signal is weak or there is no internet coverage. In situations where Wi-Fi-dependent systems would malfunction, the feeder offers a comprehensive monitoring and control solution by fusing ultrasonic sensing with GSM alerts. Additionally, the mechanical dispensing system achieved $\pm 2.67$\% consistency in portion size, outperforming traditional screw-type mechanisms \cite{amans2025design} while remaining cost-effective and energy-efficient within a low-cost IoT framework. Besides, our prototype is fully battery powered, so power outages will not hamper its functionality.

Despite these achievements, the current prototype has several limitations. Firstly, the system relies on cellular network coverage; though its availability is more common than Wi-Fi, it cannot function where there is no network signal at all. Secondly, charging the battery timely is very important to make the prototype stay running. Though our battery pack can support the prototype for more than 3 days. Future work will focus on adding solar-powered battery support for extended operation. Advanced sensors, such as weight sensors, will also be added to improve the monitoring of a variety of food types and environmental conditions. Furthermore, camera integration and pet recognition algorithms can be added to provide customized feeding for multi-pet households and remote monitoring.

\section{Conclusion}

The design, creation, and testing of a fully functional GSM-IoT smart pet feeder are described in this paper.  This provides Internet of Things capabilities that are not dependent on internet connectivity. By integrating sensing, actuation, and communication modules into a cohesive system, it demonstrates that practical IoT solutions can be built using widely available GSM networks. This approach ensures both reliable operation in remote areas and broadens the range of deployment of smart pet care system. The project fills a significant technological gap in pet care by offering an inexpensive, easily accessible Internet of Things solution that works in a variety of environmental settings.

\bibliographystyle{IEEEtran}
\bibliography{ref}

\end{document}